\documentclass{article}
\pdfoutput=1
\usepackage{microtype}
\usepackage{graphicx}
\usepackage{subfigure}
\usepackage{booktabs} 
\usepackage{amsmath,amssymb,amsfonts}

\usepackage{hyperref}

\usepackage[accepted]{icml2021}

\icmltitlerunning{A Turing Test for Transparency}

\begin{document}

\twocolumn[
\icmltitle{A Turing Test for Transparency}




\begin{icmlauthorlist}
\icmlauthor{Felix Biessmann}{bht,ecdf}
\icmlauthor{Viktor Treu}{bht}
\end{icmlauthorlist}

\icmlaffiliation{bht}{Beuth University of Applied Sciences, Berlin}
\icmlaffiliation{ecdf}{Einstein Center Digital Future, Berlin}

\icmlcorrespondingauthor{Felix Biessmann}{felix.biessmann@beuth-hochschule.de}

\icmlkeywords{Explainable AI, XAI, Interpretable Machine Learning, }

\vskip 0.3in
]



\printAffiliationsAndNotice{\icmlEqualContribution} 

\begin{abstract}
A central goal of explainable artificial intelligence (XAI) is to improve the trust relationship in human-AI interaction. One assumption underlying research in transparent AI systems is that explanations help to better assess predictions of machine learning (ML) models, for instance by enabling humans to identify wrong predictions more efficiently. Recent empirical evidence however shows that explanations can have the opposite effect: When presenting explanations of ML predictions humans often tend to trust ML predictions even when these are wrong. Experimental evidence suggests that this effect can be attributed to how intuitive, or human, an AI or explanation appears. This effect challenges the very goal of XAI and implies that responsible usage of transparent AI methods has to consider the ability of humans to distinguish machine generated from human explanations. 
Here we propose a quantitative metric for XAI methods based on Turing's imitation game, a {\em Turing Test for Transparency}. A human interrogator is asked to judge whether an explanation was generated by a human or by an XAI method. Explanations of XAI methods that can not be detected by humans above chance performance in this binary classification task are passing the test. Detecting such explanations is a requirement for assessing and calibrating the trust relationship in human-AI interaction. 
We present experimental results on a crowd-sourced text classification task demonstrating that even for basic ML models and XAI approaches most participants were not able to differentiate human from machine generated explanations. We discuss ethical and practical implications of our results for applications of transparent ML. 
\end{abstract}

\section{Introduction}
\label{sec:intro}

Machine learning (ML) systems are increasingly being used to automate or assist decision making. The growing complexity of ML systems is accompanied by an increased demand for transparency in automated decision making, both for technical and for ethical reasons. Transparency of an ML system can help to debug ML models \cite{Lapuschkin2019}. And interpretable ML can increase trust in ML technology in human-AI interaction, for instance when revealing biases of a trained model \cite{Phillips2018}. 

A common goal of XAI methods is to generate explanations that render ML predictions comprehensible to humans. The underlying assumption is that understanding ML predictions is a requirement for trusting ML predictions.   The quality of explanations can be empirically evaluated based on the concept of {\em simulatability}, meaning how helpful an explanation is for humans to replicate the ML prediction \cite{lipton2016mythos,Poursabzi-Sangdeh2018,Hase2020}. 

Empirical evaluation of XAI quality based on simulatability demonstrates that explanations often do not contribute to a well calibrated trust relationship in human-AI collaboration\footnote{Well calibrated here refers to a trust relationship in which neither blind trust, meaning humans follow (transparent) AI predictions independent of them being correct or wrong, nor ignorance of useful AI advice occurs.}: explanations often do not help to improve simulatability \cite{Hase2020} and explanations do not help to identify wrong AI predictions \cite{Poursabzi-Sangdeh2018}. 
These findings challenge some the assumptions of XAI research: If explanations do not enable humans to identify wrong predictions and do not improve interpretability which metric should XAI researchers strive to optimize? 

\begin{figure}
\centering
\includegraphics[width=.8\columnwidth]{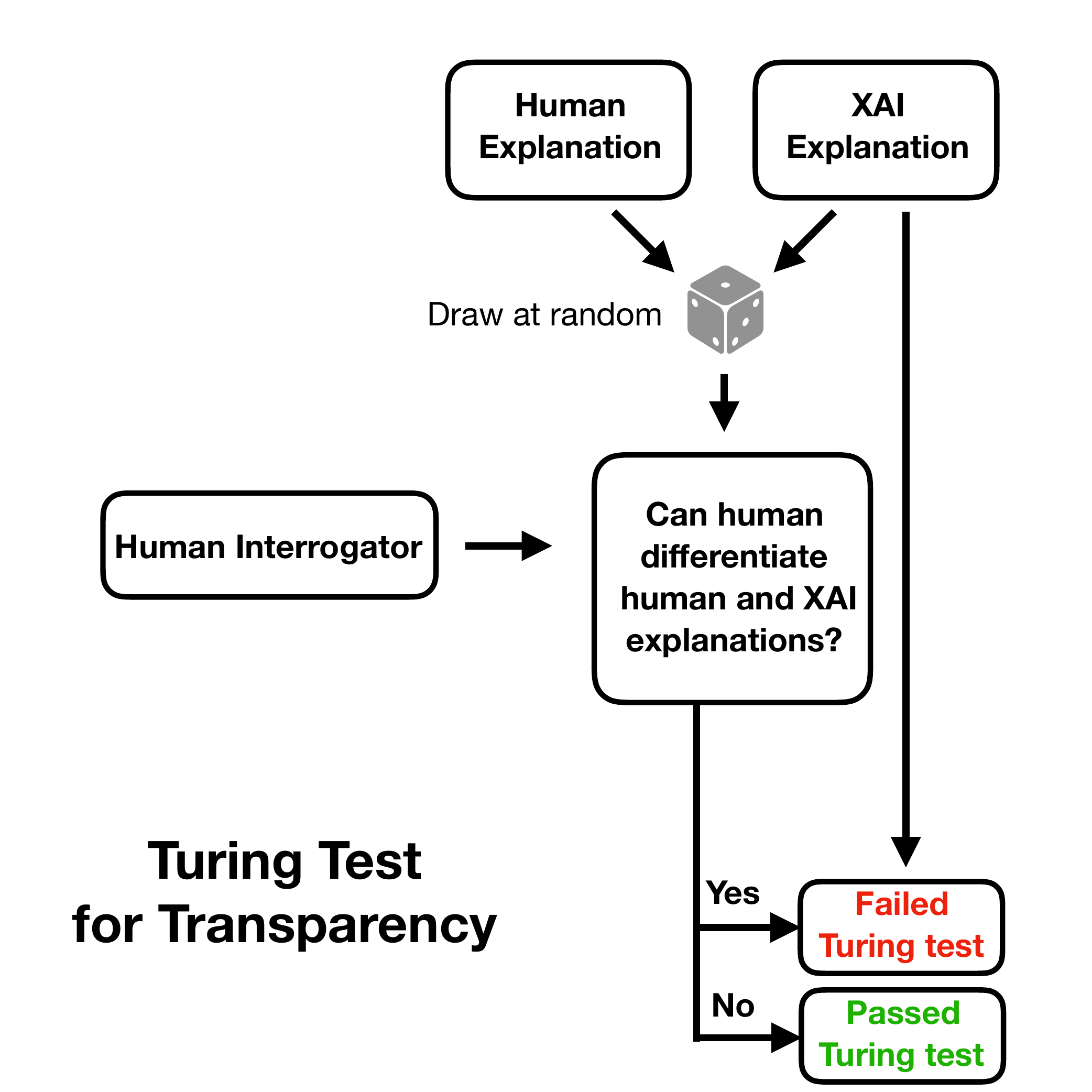}\\
\caption{\textbf{Turing Test for Transparency} proposed in this study, in analogy to Turing's imitation game. Human interrogators are asked to differentiate human from machine generated explanations in a decision making task. If humans perform at chance level, the machine generated explanations are passing the test.
}
\label{fig:ttt-setup}
\end{figure}

One observation that was made in studies investigating the human-AI trust relationship is that humans tend to trust humans more than AI systems -- even when humans are known to perform worse than the AI \cite{Dietvorst2015}. This effect is also observed in XAI studies, in which researchers find that humans would follow wrong transparent AI predictions, if the explanation appears intuitive or human \cite{Schmidt2020}. 
These findings suggest that controlling for how intuitive or human an explanation appears is an important -- and so far underrepresented -- aspect of machine generated explanations. 
Here we present, to the best of our knowledge, the first approach to measure this aspect of machine generated explanations. 
In analogy to the idea put forward by Alan Turing, which he himself referred to as the \textit{imitation game}, now referred to as the \textit{Turing Test} \cite{Turing1950}, we here propose a \textit{Turing Test for Transparency}. Similar to the original imitation game we assume that there is a human interrogator who is presented with the output of a ML model or a human. In addition to the original imitation game we also provide an explanation for the prediction made by a computer or a human subject. The task of the interrogator is now to decide whether the explanation was generated by a human or a computer. An illustration of the experiment is sketched in \autoref{fig:ttt-setup}.

\section{Experiments}
\label{sec:experiments}
We conducted two experiments, one for data collection and one for the actual Turing Test on the crowd-working platform Amazon Mechanical Turk. We recruited 145 workers, each worker participated in both experiments after passing a simple bot detection mechanism\footnote{Participants were asked what the annotation task is about and qualified as human if the subject chose the correct out of three possible answers.}. 

\subsection{Data Set, Classification Model and XAI Method}
We used a publicly available IMDb movie review sentiment dataset originally introduced in \cite{Maas2011} and selected a set of reviews for which the ML model’s classification accuracy was 80\%. We used a unigram bag-of-words feature extractor followed by term-frequency inverse document frequency normalization. The sparse feature vectors where then fed into an $L_2$ regularized Logistic Regression model and trained using stochastic gradient descent. The regularization was optimized using grid search using scikit-learn \cite{scikit-learn}. This basic text classification pipeline was trained on 25,000 movie reviews and achieved precision/recall/f1-scores of 0.87 on a test set of also 25,000 samples. Explanations for each review were computed based on covariance of unigram features with the class likelihood \cite{schmidt2019quantifying}.

\subsection{Experiment 1: Collecting Human Explanations}
\label{sec:exp-1}
In order to compare explanations of humans and the ML model we acquired five samples of human explanations and annotations for movies randomly drawn from the set of 50 movie reviews. Subjects were asked to classify the reviews as positive or negative and mark the three words most relevant for their decision. 

\subsection{Experiment 2: Turing Test for Transparency}
\label{sec:exp-2}
Each subject was shown five movie reviews, again drawn at random from those reviews they have not seen in the first experiment. In contrast to the first experiment subjects were also shown the prediction of a human or the ML model along with the respective explanation for the prediction. 
Both human and machine generated explanations were represented by highlighting the three words most relevant for the prediction.

\section{Results}
\label{sec:results}
After all 145 subjects performed both experiments we filtered out participants who did not achieve an annotation accuracy of at least 60\%, which reduced the number of subjects to 133. 
\begin{table*}
\centering
\caption{Comparison of explanations generated by human subjects and AI. Shown are five words for each set (selected by humans and not AI, selected by AI but not by humans and selected by both humans and AI), randomly drawn for positive and negative class after controlling for the number of words in human and AI explanations. Qualitative comparison suggests that the explanations generated by humans and the XAI method used are similar with respect to semantic and syntactic features.  }
\label{tab:explanations-exp-1}
\vspace{1em}
\begin{tabular}{lllllll}
\toprule
        \multicolumn{3}{c}{positive} &   \multicolumn{3}{c}{negative}\\
        \midrule
Humans $\setminus$ AI &  AI $\setminus$ Human & Humans $\cap$ AI &Humans $\setminus$ AI &  AI $\setminus$ Human & Humans $\cap$ AI\\
      \midrule
\parbox[c]{2.4cm}{extraordinary, impressed, recommend, hilarious,  interestingly} &
\parbox[c]{2.4cm}{human, family, story,  lovable, relationship} & 
\parbox[c]{2.4cm}{magnificent, brilliant, enjoyed, excellent, superb} & 
\parbox[c]{2.4cm}{nonsensical, unfunny, disappointed, recommended, sick} &
\parbox[c]{2.4cm}{disappointing, avoid, thing, reason, premise} &
\parbox[c]{2.4cm}{dull, horrible, ridiculous, worst, awful} \\
\bottomrule
\end{tabular}
\end{table*}

\begin{table}
\caption{Results of Turing Test for Transparency. Shown are precision, recall, F1 score and accuracy for the task of distinguishing human generated explanations from machine generated explanations. All metrics indicate that humans were not able to distinguish human and AI explanations. This can be interpreted as the machine generated explanations passing the Turing Test.}
\label{tab:results-exp-2}
\vspace{1em}
\begin{tabular}{lrrrr}
\toprule
{} &  precision &  recall &  f1-score &  support \\
\midrule
ML model          &  0.49 &  0.50 &  0.49 &  311 \\
human        &  0.56 &  0.56 &  0.56 &  359 \\
accuracy     &  0.53 &  0.53 &  0.53 &  \\
weighted avg &  0.53 &  0.53 &  0.53 &  670\\
\bottomrule
\end{tabular}
\end{table}

\subsection{Results Turing Test for Transparency}
\label{sec:results-ttt}
We first quantified whether humans were able to distinguish human generated from AI generated explanations. Note that we discarded those human generated explanations obtained in experiment 1 for which human annotations were incorrect. The results are shown in \autoref{tab:results-exp-2} as standard classification metrics, precision, recall, F1 score for both human and AI generated explanations and accuracy, aggregated across both classes. 

\begin{figure}
\includegraphics[width=\columnwidth]{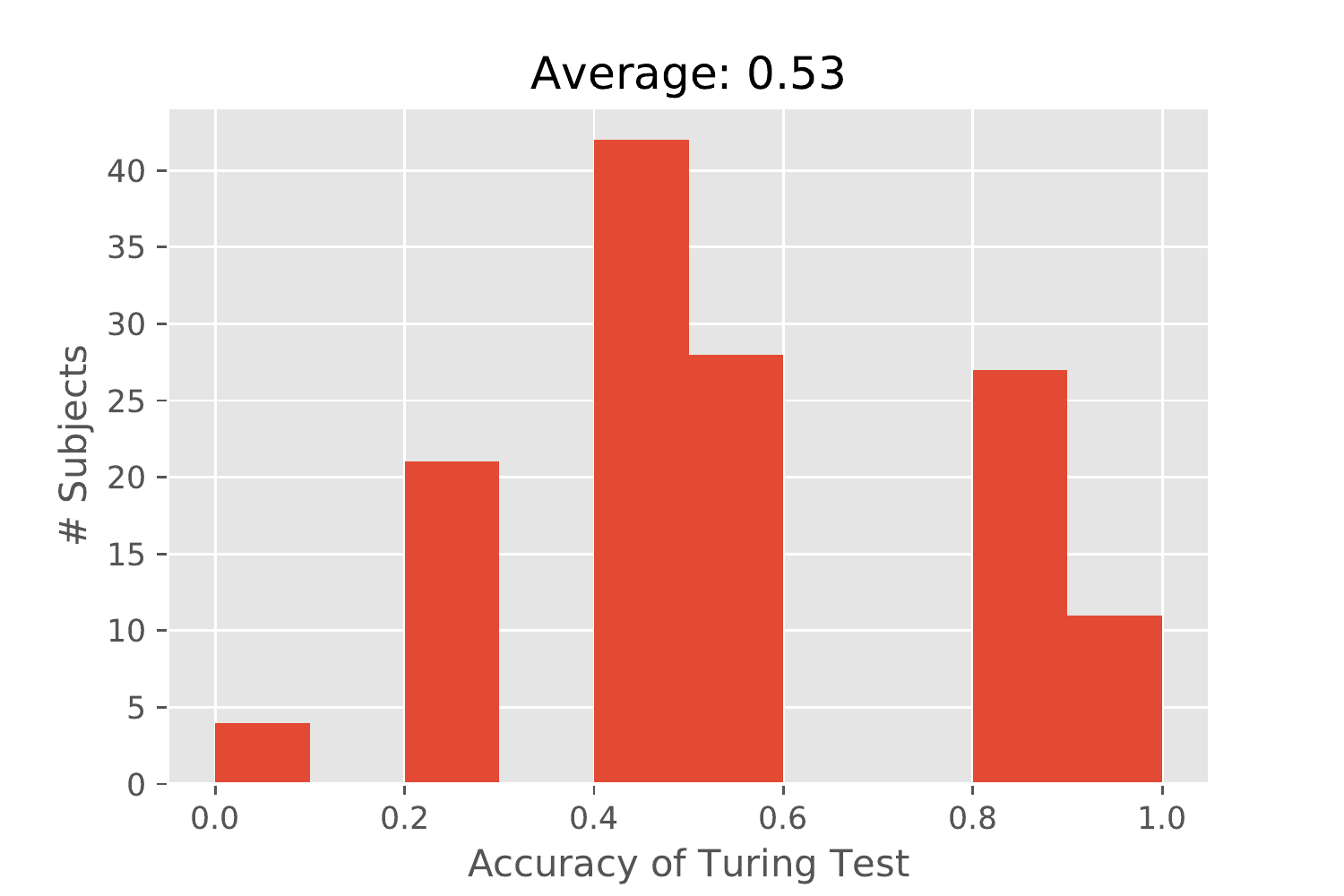}\\
\caption{Results of turing test for transparency. Shown are histogram of accuracies with which human subjects can classify correctly human vs. AI explanations. On average humans performed around chance with an overall accuracy of 53\%, indicating that the explanations provided by the AI were passing the turing test.}
\label{fig:turing-test-results}
\end{figure}

\paragraph{XAI Explanations pass Turing Test}
All metrics demonstrate that in this particular experiment humans were not able to differentiate human from AI explanations as the metrics indicate that human annotators performed around chance performance in the Turing Test.

We further investigated the distribution of accuracies in the Turing Test across subjects to assess whether there is any systematic pattern or bias in the set of participants sampled. In \autoref{fig:turing-test-results} we show the distribution of accuracies in the Turing Test across all subjects. There seems to be no particular bias in the subject sample and the overall distribution appears to be close to a normal distribution centered around 0.53. 

\paragraph{Annotation accuracy vs. Turing Test accuracy: Grouped by subjects}
One might hypothesize that the variability across participants to differentiate human from XAI explanations is due to cognitive abilities that also enable subjects to perform well in the annotation task itself. In order to investigate this relationship computed the correlation between the annotation accuracy of each annotator and their performance in the Turing Test. The correlation was 0.14, indicating that there is no strong dependency between task performance and Turing Test performance. This effect could be explained by different cognitive mechanisms governing human decisions in the two tasks. 

\begin{figure}
\includegraphics[width=\columnwidth]{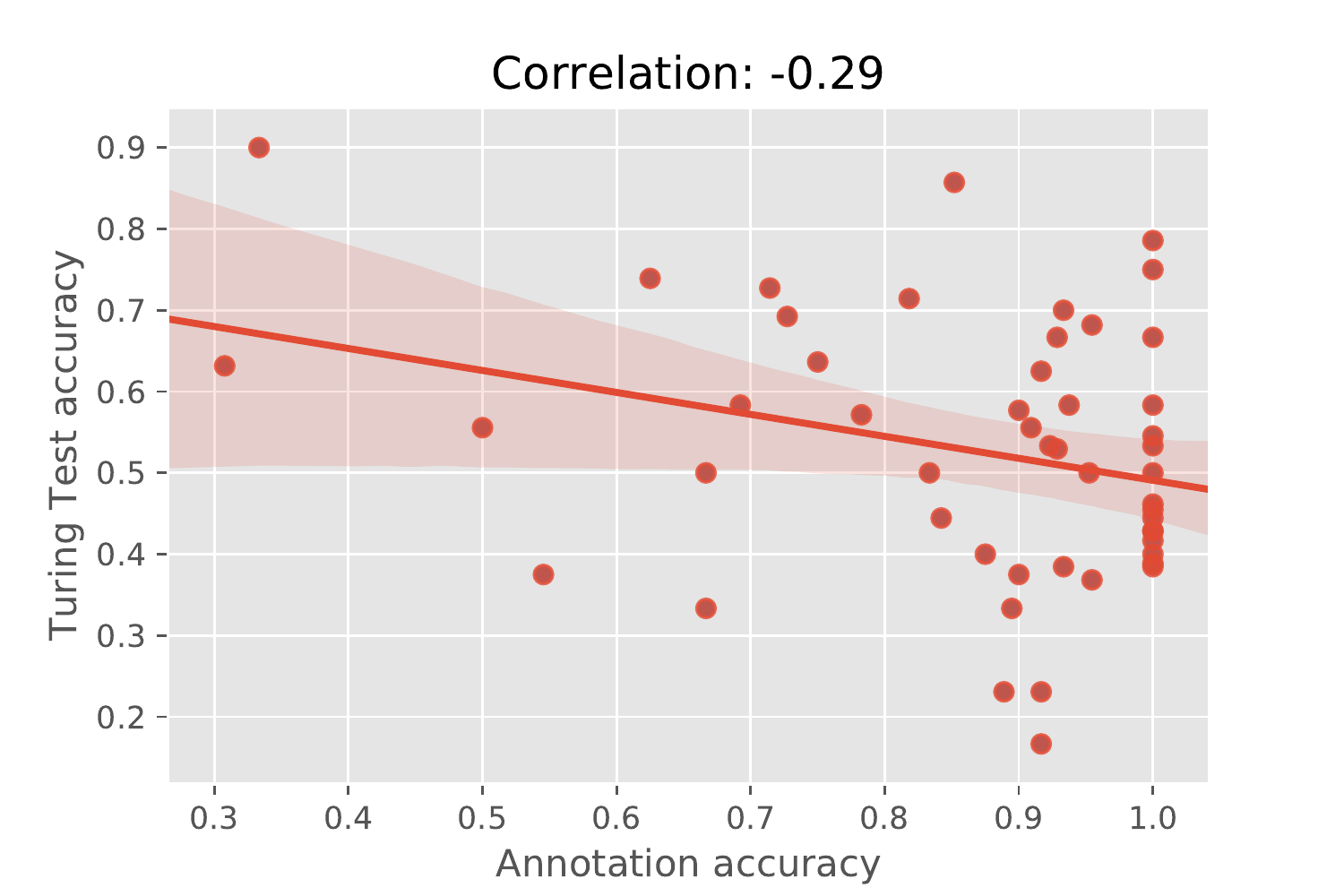}\\
\caption{The accuracy of human annotators in the binary movie review annotation task shows a slightly negative correlation with the accuracy of human annotators in the Turing Test for Transparency. Explanations corresponding to movie reviews that were correctly annotated were more difficult to distinguish from machine generated explanations. Note that human generated explanations for reviews that were incorrectly annotated by humans were discarded. }
\label{fig:accuracy_tt_vs_annotation}
\end{figure}

\paragraph{Annotation accuracy vs. Turing Test accuracy: Grouped by reviews}
When grouping the data by movie reviews we see a slightly stronger but negative correlation of -0.29, as shown in \autoref{fig:accuracy_tt_vs_annotation}. This trend indicates that human generated explanations for reviews that were often annotated correctly in the original classification task tended to fail the Turing Test for Transparency more often. In contrast human generated explanations movie reviews which were often annotated incorrectly were often easier to differentiate from machine generated explanations. Note that all human generated explanations for which human annotators provided a wrong annotation were discarded from the Turing Test for Transparency. 

\begin{figure}
\includegraphics[width=\columnwidth]{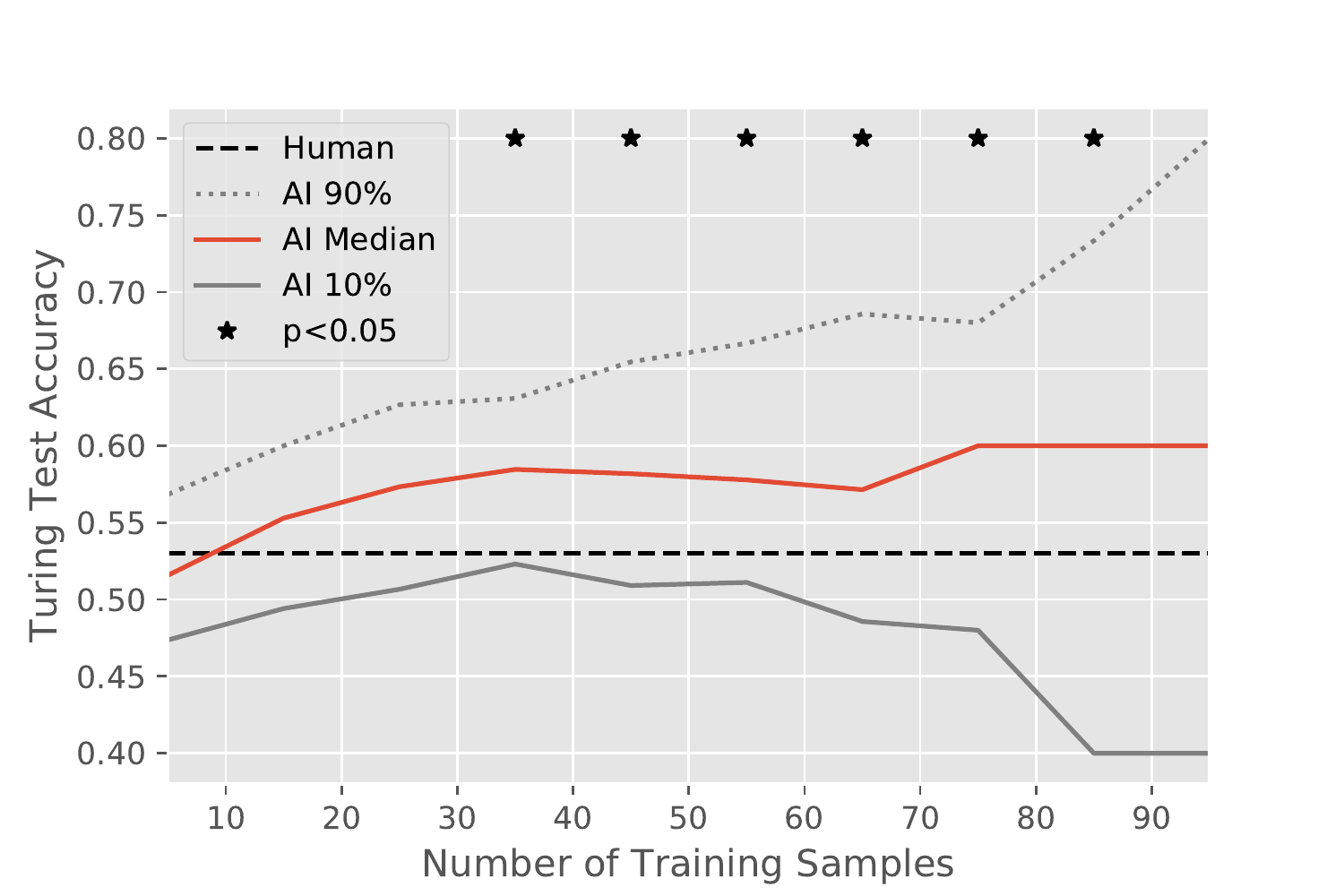}\\
\caption{Accuracy of passing the Turing Test by an AI or human subjects. Shown are the human accuracy \textit{black dashed line}) and the accuracy (10th/50th/90th quantile) of an AI for varying training set size (shown on \textit{x-axis}). Stars indicate the training set sizes for which the accuracy distribution over ML models was significantly different from the Turing Test accuracy distribution across all human subjects ($p<0.05$, Bonferroni corrected).}
\label{fig:turing-test-human-vs-ai}
\end{figure}

\paragraph{Turing Test Accuracy: Humans worse than AI}
We trained the same model used for the original text classification task on the binary classification task of the Turing Test for Transparency. 
For each movie review the human or machine generated explanation was the input to the ML model and the binary label was \textit{human generated} or \textit{machine generated}. 
Model training was done for different training set sizes to estimate the sample efficiency of the model. In \autoref{fig:turing-test-human-vs-ai} we show the human accuracy as a baseline and the 10th/50th/90th quantile of a set of ML models trained on equally sized independently sampled training data sets. The number of ML models trained for each training set size was chosen to match the number of subjects. 

Our results show that for small training set sizes the ML models did not achieve accuracies significantly larger than those of human subjects. 
For larger training set sizes however the ML model appeared to achieve slightly but significantly higher accuracies compared to humans. We tested for significance using a Kruskal-Wallis test for equal variances across the two distributions of human and ML model accuracies and corrected for multiple comparisons using Bonferroni correction. 
These results show that in this particular example, ML models can better differentiate human from machine generated explanations when the training set size was larger than 30 reviews. 
One could argue that the ML models only achieve significantly higher accuracies when seeing more training data than human subjects did in our experiment. But on the other hand the implicit assumption of the Turing Test is that humans have seen so many training examples, at least of the human generated class, that they know how human explanations look like. 

\section{Conclusion}
\label{sec:conclusion}

A common assumption of XAI research is that explanations increase trust in human-AI collaboration by making it easier to spot wrong predictions \cite{Lapuschkin2019} and to speed up and improve the confirmation of correct ML predictions \cite{schmidt2019quantifying}. These assumptions are challenged by increasing evidence for XAI methods leading to biases in human-AI interation, for example when humans trust transparent ML predictions even when they are wrong \cite{Poursabzi-Sangdeh2018,Schmidt2020}. One reason for this bias could be explained by the empirical observation that humans tend to trust ML systems less than humans, even when they know that the ML system performs better \cite{Dietvorst2015}. 
These findings suggest that in order to calibrate the XAI system to the right level of transparency, it is important to consider how intuitive or human an explanation is. Here we present a quantitative metric in analogy to the imitation game (Turing Test) that directly captures this effect. 
Our results demonstrate that some machine generated textual explanations cannot be differentiated from human generated explanations. Future work will explore this effect with more complex explanations.
While this can be regarded as an achievement of XAI research, it also raises ethical questions on the application of transparent ML methods. If humans are biased towards blindly following intuitive explanations we argue that considering humans' ability to detect machine generated explanations is an important factor for responsible usage of XAI methods.  We hope that the proposed XAI quality metric can contribute to a better quantitative evaluation of transparent ML methods.

\bibliography{references}
\bibliographystyle{icml2021}

\end{document}